\documentclass[runningheads]{llncs}
\usepackage{graphicx}

\usepackage{times}
\usepackage{url}
\usepackage{latexsym}

\usepackage{microtype}
\usepackage{booktabs}
\usepackage{makecell}
\usepackage{tabularx}
\usepackage{array}
\newcolumntype{C}{>{\centering\arraybackslash}p{4.7em}}
\usepackage{multirow}
\usepackage{colortbl}
\usepackage{amsmath}
\usepackage{amsfonts}
\usepackage{mathabx}
\usepackage{graphicx}
\usepackage{subcaption}
\usepackage{url}

\usepackage{caption}
\usepackage{arydshln}
\usepackage{todonotes}

\begin{document}
%
%\title{Evaluating Contextual Representation Models in Unsupervised Retrieval}
\title{Evaluating Multilingual Text Encoders for Unsupervised Cross-Lingual Retrieval}

\author{Robert Litschko\inst{1} \and
Ivan Vuli\'{c}\inst{2} \and
Simone Paolo Ponzetto\inst{1} \and
Goran Glava\v{s}\inst{1}}
\authorrunning{R. Litschko et al.}
% First names are abbreviated in the running head.
% If there are more than two authors, 'et al.' is used.
%
\institute{Data and Web Science Group, University of Mannheim, Mannheim, Germany \\
\email{\{litschko,simone,goran\}@informatik.uni-mannheim.de} \and
Language Technology Lab, University of Cambridge, UK \\
\email{iv250@cam.ac.uk}}

\maketitle              % typeset the header of the contribution
\begin{abstract}
Pretrained multilingual text encoders based on neural Transformer architectures, such as multilingual BERT (mBERT) and XLM, have achieved strong performance on a myriad of supervised language understanding tasks. Consequently, they have been adopted as a state-of-the-art paradigm for multilingual and cross-lingual representation learning and transfer, rendering cross-lingual word embeddings (CLWEs) effectively obsolete. However, questions remain to which extent this finding generalizes 1) to unsupervised settings and 2) for ad-hoc cross-lingual IR (CLIR) tasks. Therefore, in this work we present a systematic empirical study focused on the suitability of the state-of-the-art multilingual encoders for cross-lingual document and sentence retrieval tasks across a large number of language pairs. In contrast to supervised language understanding, our results indicate that for unsupervised document-level CLIR -- a setup in which there are no relevance judgments for task-specific fine-tuning -- the pretrained encoders fail to significantly outperform models based on CLWEs. For sentence-level CLIR, we demonstrate that state-of-the-art performance can be achieved. However, the peak performance is not met using the general-purpose multilingual text encoders `off-the-shelf', but rather relying on their variants that have been further specialized for sentence understanding tasks.


%Multilingual pretrained encoders have been extensively used in language processing and yield state-of-the-art performance. In this work we benchmark multilingual BERT (mBERT) and XLM on document- and sentence-level unsupervised cross-lingual retrieval (CLIR). We show that the models, out of the box, yield mixed results compared to bilingual word embeddings (BLWE). Recent work specialize encoders to produce meaningful sentence representations for improved zero-shot applications. We benchmark a range of those models and show that they yield SOTA performance on CLIR, by a large margin on sentence-level CLIR and by a considerate margin on document-level CLIR. 
%\keywords{First keyword  \and Second keyword \and Another keyword.}
%% (e.g., natural language inference question answering)

\end{abstract}

\section{Introduction}
Cross-lingual information retrieval (CLIR) systems respond to queries in a source language by retrieving relevant documents in another, target language. Their success is typically hindered by data scarcity: they operate in challenging low-resource settings without sufficient labeled training data, i.e., human relevance judgments, to build supervised models (e.g., neural matching models for pair-wise retrieval \cite{10.1145/3397271.3401322,jiang2020cross}). This motivates the need for robust, resource-lean and unsupervised CLIR approaches.

In previous work, Litschko et al.\ \cite{litschko2019evaluating} have shown that language transfer through cross-lingual embedding spaces (CLWEs) can be used to yield state-of-the-art performance in a range of unsupervised ad-hoc CLIR setups. This approach uses very weak supervision (i.e., only a bilingual dictionary spanning 1K-5K word translation pairs), or even no supervision at all, in order to learn a mapping that aligns two monolingual word embedding spaces \cite{glavas-etal-2019-properly,Vulic:2019emnlp}. Put simply, this enables casting CLIR tasks as 'monolingual tasks in the shared (CLWE) space': at retrieval time both queries and documents are represented as simple aggregates of their constituent CLWEs. However, this approach, by limitations of static CLWEs, cannot capture and handle polysemy in the underlying text representations. \textit{Contextual text representation models} alleviate this issue \cite{liu2020survey}. They encode occurrences of the same word differently depending on its surrounding context.

%allow  words to have different representations across different contexts.
%%  since when using CLWEs multiple meanings are squeezed into a single representation.

Such contextual representations are obtained via large models pretrained on large text collections through general objectives such as (masked) language modeling \cite{mbert,Liu:2019roberta}. Multilingual text encoders pretrained on 100+ languages, such as mBERT \cite{mbert} or XLM \cite{xlm}, have become a \textit{de facto} standard for multilingual representation learning and cross-lingual transfer in natural language processing (NLP). These models demonstrate state-of-the-art performance in a wide range of supervised language understanding and language generation tasks \cite{xcopa,xglue}, especially in zero-shot settings: a typical \textit{modus operandi} is fine-tuning a pretrained multilingual encoder with task-specific data of a source language (typically English) and then using it directly in a target language. 

However, there are still several crucial questions remaining. First, it is unclear whether these general-purpose multilingual text encoders can be used directly for ad-hoc CLIR without any additional supervision (i.e., relevance judgments). Further, can they outperform the previous unsupervised CLIR approaches based on static CLWEs \cite{litschko2019evaluating}? How do they perform depending on the (properties of the) language pair at hand? How can we encode useful semantic information using these models, and do different ``encoding variants'' (see later \S\ref{s:mencoders}) yield different retrieval results? Are there performance differences in unsupervised sentence-level versus document-level CLIR tasks? Finally, can we boost performance by relying on sentence encoders that are specialized towards dealing with sentence-level understanding in particular? In order to address these questions, we present a systematic empirical study and profile the suitability of state-of-the-art pretrained multilingual encoders for different CLIR tasks and diverse language pairs. We evaluate two state-of-the-art general-purpose pretrained multilingual encoders, mBERT \cite{mbert} and XLM \cite{xlm} with a range of encoding variants, and also compare them to CLIR approaches based on static CLWEs, and specialized multilingual sentence encoders. Our key contributions can be summarized as follows:

% In multilingual encoders language transfer is facilitated by using a shared word-piece vocabulary and training the model on massive amounts of multilingual text.
%There exists many pre-trained models that can be used out of the box for cross-lingual fine-tuning: here, we investigate two popular models, mBERT and XLM, on CLEF document-level retrieval and Europarl sentence-level retrieval. Moreover, with the rising interest in zero-shot or few-shot learning, i.e., training a multilingual encoder on a resource-rich language and applying it on a resource-scarce language with little to no extra training, much research has focused on developing general-purpose multilingual sentence encoders \cite{artetxe2019LASER,yang-etal-2020-multilingual-muse,feng2020language-labse,reimers2020making-distil}. Consequently, besides mBERT and XLM, we  additionally benchmark multilingual sentence encoders in our unsupervised CLIR setup. Generally, our overarching aim with this paper is to show how the original findings Litschko et al.\ \cite{litschko2019evaluating} transfer and compare with contextualized multilingual word embeddings -- one would expect these new kids on the block among text representations to always yield better results \emph{tout court}, but is this really the case? Our contributions are as follows:

%\begin{enumerate}
\vspace{1.4mm}
\noindent \textbf{(1)} We empirically validate that, without any task-specific fine-tuning, multilingual encoders such as mBERT and XLM fail to outperform CLIR approaches based on static CLWEs. Their performance also crucially depends on how one encodes semantic information with the models (e.g., treating them as sentence/document encoders directly versus averaging over constituent words and/or subwords). We also show that there is no ``one-size-fits-all'' approach, and the results are task- and language-pair-dependent.

\vspace{1.3mm}
\noindent \textbf{(2)} We provide a first large-scale comparative evaluation of state-of-the art pretrained multilingual encoders on unsupervised document-level CLIR. We also empirically show that encoder models specialized for sentence-level understanding substantially outperform general-purpose models (mBERT and XLM) on sentence-level CLIR tasks.
%\end{enumerate}

\section{Related Work}
\label{sec:rw}

%\vspace{1em}
\noindent
\textbf{Self-Supervised Pretraining and Transfer Learning.} Recently, research on universal sentence representations and transfer learning has gained much traction. InferSent \cite{conneau-EtAl:2017:EMNLP2017} transfers the encoder of a model trained on natural language inference to other tasks, while USE \cite{cer2018universal} extends this idea to a multi-task learning setting. More recent work explores self-supervised neural Transformer-based \cite{vaswani2017attention} models based on (causal or masked) language modeling (LM) objectives such as BERT \cite{mbert}, RoBERTa \cite{Liu:2019roberta}, GPT \cite{radford2019language,brown2020language}, and XLM \cite{xlm}.\footnote{Note that self-supervised learning can come in different flavors depending on the training objective \cite{Clark:2020iclr}, but language modeling objectives still seem to be the most popular choice.} Results on benchmarks such as GLUE \cite{wang2018glue} and SentEval \cite{conneau2018senteval} indicate that these models can yield impressive (sometimes human-level) performance in supervised Natural Language Understanding (NLU) and Generation (NLG) tasks. These models have become \emph{de facto} standard and omnipresent text representation models in NLP. In supervised monolingual IR, self-supervised LMs have been employed as contextualized word encoders \cite{macavaney2019cedr}, or fine-tuned as pointwise and pairwise rankers \cite{nogueira2019multi}.
%
% Skip-thought \cite{kiros2015skip} uses an encoding model to produce sentence representations in order to predict the token sequences of the preceding and following sentence. 
%
%% For example, in Masked Language Modelling (MLM) [13] the model is given a corruptsentence, each token is with a certain probability either replaced with a random token ora [MASK] token from which the model has to reconstruct the original sentence. 

\vspace{1.8mm}
\noindent
\textbf{Multilingual Text Encoders} based on the (masked) LM objectives have also been massively adopted in multilingual and cross-lingual NLP and IR applications. A multilingual extension of BERT (mBERT) is trained with a shared subword vocabulary on a single multilingual corpus obtained as concatenation of large monolingual data in 104 languages. The XLM model \cite{xlm} extends this idea and proposes natively cross-lingual LM pretraining, combining causal language modeling (CLM) and translation language modeling (TLM).\footnote{In CLM, the model is trained to predict the probability of a word given the previous words in a sentence. TLM is a cross-lingual variant of standard masked LM (MLM), with the core difference that the model is given pairs of parallel sentences and allowed to attend to the aligned sentence when reconstructing a word in the current sentence.}  Strong performance of these models in supervised settings is confirmed across a range of tasks on multilingual benchmarks such as XGLUE \cite{xglue} and XNLI \cite{conneau2018xnli}. However, recent work \cite{reimers2020making-distil,cao2019multilingual} has indicated that these general-purpose models do not yield strong results when used as out-of-the-box text encoders in an unsupervised transfer learning setup. We further investigate these preliminaries, and confirm this finding also for unsupervised ad-hoc CLIR tasks. 

Multilingual text encoders have already found applications in document-level CLIR. Jiang et al. \cite{jiang2020cross} use mBERT as a matching model by feeding pairs of English queries and foreign language documents. MacAvaney et al. \cite{macavaney2020teaching} use mBERT in a zero-shot setting, where they train a retrieval model on top of mBERT on English relevance data and apply it on a different language. However, prior work has not investigated unsupervised CLIR setups, and a systematic comparative study focused on the suitability of the multilingual text encoders for diverse ad-hoc CLIR tasks and language pairs is still lacking.

%%\todo{TODO: Need to better say how contribution is different from these papers.}

%% IV: I don't understand what this sentence is trying to say and how/where this fits:
%% "Applications such as bitext mining have gained interest because they are useful for augmenting training data for resource-hungry NMT models, or indexing question-answer pairs for automated FAQ answer retrieval."

\vspace{1.8mm}
\noindent
\textbf{Specialized Multilingual Sentence Encoders.}
% The general theme is that universal sentence encoders are either obtained by utilizing ideas from machine translation \cite{artetxe2019LASER,mccann2017learned}, or by distilling compositional sentence knowledge from a monolingual model into a self-supervised multilingual model \cite{reimers2019sentence,feng2020language-labse,yang-etal-2020-multilingual-muse}. For example, in
An extensive body of work focuses on inducing multilingual encoders that capture sentence meaning. In \cite{artetxe2019LASER}, the multilingual encoder of a sequence-to-sequence model is shared across languages and optimized to be language-agnostic, whereas Guo et al. \cite{guo-etal-2018-effective} rely on a dual Transformer-based encoder architectures instead (with tied/shared parameters) to represent parallel sentences. Rather than optimizing for translation performance directly, their approach minimizes the cosine distance between parallel sentences. A ranking softmax loss is used to classify the correct (i.e., aligned) sentence in the other language from negative samples (i.e., non-aligned sentences). In \cite{yang2019improving}, this approach is extended by using a bidirectional dual encoder and adding an additive margin softmax function, which serves to push away non-translation-pairs in the shared embedding space. The dual-encoder approach is now widely adopted  \cite{guo-etal-2018-effective,yang-etal-2020-multilingual-muse,feng2020language-labse,reimers2020making-distil,zhao2020inducing}, and yields state-of-the-art multilingual sentence encoders which excel in sentence-level NLU tasks.

Other recent approaches propose input space normalization, and re-aligning mBERT and XLM with parallel data \cite{zhao2020inducing,cao2019multilingual}, or using a teacher-student framework where a student model is trained to imitate the output of the teacher network while preserving high similarity of translation pairs \cite{reimers2020making-distil}. In \cite{yang-etal-2020-multilingual-muse}, authors combine multi-task learning with a translation bridging task to train a universal sentence encoder. We benchmark a series of representative sentence encoders; their brief descriptions are provided in \S\ref{sec:sent-specialized-models}.

%%While most of the existing approaches focus on minimizing cosine distance between parallel text the authors of \cite{feng2020language-labse} re-use MLM and TLM during fine-tuning. 

\vspace{1.8mm}
\noindent
\textbf{CLIR Evaluation and Application.} The cross-lingual ability of mBERT and XLM has been investigated by probing and analyzing their internals \cite{karthikeyan2019cross}, as well as in terms of downstream performance \cite{pires2019multilingual,wu2019beto}. In CLIR, these models as well as dedicated multilingual sentence encoders have been evaluated on tasks such as cross-lingual question-answer retrieval \cite{yang-etal-2020-multilingual-muse}, bitext mining \cite{ziemski2016united,ZWEIGENBAUM18.12-BUCC}, and semantic textual similarity (STS) \cite{hoogeveen2015,lei2016semi}. Yet, the models have been primarily evaluated on sentence-level retrieval, while classic ad-hoc (unsupervised) document-level CLIR has not been in focus. Further, previous work has not provided a large-scale comparative study across diverse language pairs and with different model variants, nor has tried to understand and analyze the differences between sentence-level and document-level tasks. In this work, we aim to fill these gaps. 

%% (IV, not really important): Especially mining parallel text is useful for augmenting training data of resource-hungry NMT models.

%% (IV, not really important):  Question-Answer-pair retrieval is motivated by the application of automated FAQ answer retrieval, where the closest question is identified and its associated answer is returned. 

%% (IV, I don't understand what is this trying to convey): The authors of \cite{artetxe2019LASER} criticize the small language coverage of existing benchmarks and publish a large-scale test set for similarity search on parallel sentences including low-resource languages.

% Some of the statements rephrased from
% https://slideslive.com/38928621/multilingual-universal-sentence-encoder-for-semantic-retrieval
% and
% https://ai.googleblog.com/2019/07/multilingual-universal-sentence-encoder.html
%\paragraph{}   

% Later a range of sentence encoder models have been developed \cite{yang-etal-2020-multilingual-muse,artetxe2019LASER,feng2020language-labse}. These have identified that using e.g. BERT as a scoring function of a query-document pair much more expensive than pre-computing document vectors and ranking them by cosine similarity. Remarkably multilingual language models have shown to extend to even unseen languages and languages with different skripts \cite{artetxe2019LASER,K2020Cross-Lingual}. 

%\paragraph{}
%On the level of cross-lingual IR, people have already successfully employed BERT in 

%\section{Unsupervised Cross-lingual Retrieval}
\section{Multilingual Text Encoders}
\label{s:mencoders}
We now provide an overview of all the multilingual models in our comparison. For completeness, we first  briefly describe static CLWEs (\S\ref{sec:clwes}). We then discuss mBERT and XLM as representative general-purpose multilingual text encoders trained with LM objectives (\S\ref{sec:mbert-xlm}), as well as specialized multilingual sentence encoders later in \S\ref{sec:sent-specialized-models}.

\subsection{CLIR with (Static) Cross-lingual Word Embeddings}
\label{sec:clwes}

In a standard CLIR setup, we assume a query $Q_{L_1}$ issued in a source language $L_1$, and a document collection comprising $N$ documents $D_{i, L_2}$, $i=1,\ldots,N$ in a target language $L_2$. Let $d=\{t_1,t_2,\dots,t_{|D|}\} \in D$ be a document consisting of $|D|$ terms $t_i$. A typical approach to CLIR with static CLWEs is to represent queries and documents as vectors $\overrightarrow{Q},\overrightarrow{D}\in \mathbb{R}^d$ in a $d$-dimensional shared embedding space \cite{vulic2015sigir,litschko2019evaluating}. Each term is represented independently and obtained by performing a lookup on a pre-computed static embedding table $\overrightarrow{t_i} = emb\left(t_i\right)$. There exist a range of methods for inducing shared embedding spaces with different levels of supervision, such as parallel sentences, comparable documents, small bilingual dictionaries, or even methods without any supervision \cite{ruder2019survey}. Given the shared CLWE space, both query and document representations are obtained as aggregations of their term embeddings. We follow Litschko et al. \cite{litschko2019evaluating} and represent documents as the weighted sum of their terms' vectors, where each term's weight corresponds to its inverse document frequency (idf)
%\footnote{Only document term embeddings are idf-scaled}%
: $\overrightarrow{d} = \sum_{i = 1}^{N_d}{\mathit{idf}(t^d_i) \cdot \overrightarrow{t^d_i}}$. During retrieval documents are ranked according to the cosine similarity to the query.

%% (IV, not needed now)
%%We distinguish multilingual text encoders according to the level of supervision with which they are pre-trained.

\subsection{Multilingual (Transformer-Based) Language Models: mBERT and XLM}
\label{sec:mbert-xlm}

Massively multilingual pretrained neural language models such as mBERT and XLM can be used as a dynamic embedding layer to produce contextualized word representations, since they share a common input space on the subword level (e.g. word-pieces, byte-pair-encodings) across all languages. Let us assume that a term (i.e., a word-level token) is tokenized into a sequence of $K$ subword tokens ($K\geq 1$; for simplicity, we assume that the subwords are word-pieces (\textit{wp})): $t_i=\big\{\textit{wp}_{i,k}\big\}^{K}_{k=1}$. The  multilingual encoder then produces contextualized subword embeddings for the term's $K$ constituent subwords $\overrightarrow{wp_{i,k}}$, $k=1,\ldots,K$, and we can aggregate these subword embeddings to obtain the representation of the term $t_i$: $\overrightarrow{t_i} = \psi\left(\{\overrightarrow{wp_{i,k}}\}^K_{k = 1}\right)$, where the function $\psi()$ is the aggregation function over the $K$ constituent subword embeddings. Once these term embeddings $\overrightarrow{t_i}$ are obtained, we follow the same CLIR setup as with CLWEs in \S\ref{sec:clwes}.
%
%\begin{align*}
    %enc\left(d\right)&= \Big\{ \overrightarrow{\text{WP}}_1, \overrightarrow{\text{WP}}_2, \dots, \overrightarrow{\text{WP}}_N \Big\} \\
    
    %&= \Big\{\textit{aggr}\left(\overrightarrow{\textit{wp}}_1^{(1)}\right), \dots,  \textit{aggr}\left(\overrightarrow{\textit{wp}}_{|D|}^{(K)}\right)\Big\}_i
%\end{align*}
%
%where the function $\psi(\overrightarrow{\text{WP}}_i)$ aggregates groups of sub-word embeddings $\overrightarrow{\text{WP}}_i$ belonging to the same word into single contextualized word representations $\overrightarrow{t_i}$. From here we follow the same setup as with CLWEs.

\vspace{1.8mm}

\noindent \textbf{Static Word Embeddings from Multilingual Transformers.} We first use multilingual transformers (mBERT and XLM) in two different ways to induce static word embedding spaces for all languages. In a simpler variant, we feed terms into the encoders \textit{in isolation} (\textbf{ISO}), that is, without providing any surrounding context for the terms. This effectively constructs a static word embedding table similar to what is done in \S\ref{sec:clwes}, and allows the CLIR model (\S\ref{sec:clwes}) to operate at a non-contextual word level. 
An empirical CLIR comparison between ISO and CLIR operating on CLWEs \cite{litschko2019evaluating} then effectively quantifies how well multilingual encoders (mBERT and XLM) encode word-level representations. %(i.e., lexical semantics). 

%This is useful for investigating the lexical quality of word embeddings, fully ignoring contextual knowledge. 

In a more elaborate variant we do leverage the contexts in which the terms appear, and construct \textit{average-over-contexts} embeddings (\textbf{AOC}). For each term $t$ we collect a number of sentences $s_i \in \mathcal{S}_t$ in which the term occurs. We use the full set of Wikipedia sentences $\mathcal{S}$ to sample sets of contexts $\mathcal{S}_t$ for vocabulary terms. For a given sentence $s_i$ let $j$ denote the position of $t$'s first occurrence. We then transform $s_i$ with mBERT or XLM as the encoder, $enc(s_i)$, and extract the contextualized embedding of $t$ via \textit{mean-pooling}, i.e., by averaging embeddings of its constituent subwords, $\psi\left(\{\overrightarrow{wp_{j,k}}\}^K_{k = 1}\right) = 1/K \cdot \sum_{k = 1}^{K}{\overrightarrow{wp_{j,k}}}$. 
%Here the function $\psi()$ is implemented as \textit{mean-pooling}, i.e., we obtain the contextual representation of the term as the average of contextualized vectors of its constituent subwords.
For each vocabulary term, we obtain $N_t = min(|\mathcal{S}_t|,\tau)$ contextualized vectors, with $|\mathcal{S}_t|$ as the number of Wikipedia sentences containing $t$ and $\tau$ as the maximal number of sentence samples for a term. The final static embedding of $t$ is then simply the average over the $N_t$ contextualized vectors. 
%sent  Processing each sentence $s_i$ leads to $n_s = $ contextualized representations for term $t$, where $\tau$ is the upper limit of the number of sentences considered. The final static representation $\overrightarrow{t_i}$ is obtained by taking the average of the $n_s$ extracted contextualized representations. 

The obtained static AOC and ISO embeddings, despite being induced with multilingual encoders, however, did not appear to be well-aligned across languages \cite{Liu:2019conll,cao2019multilingual}. We evaluated the static ISO and AOC embeddings induced for different languages with multilingual encoders (mBERT and XLM), on the bilingual lexicon induction (BLI) task \cite{glavas-etal-2019-properly}. We observed poor BLI performance, suggesting that further projection-based alignment of respective monolingual ISO and AOC spaces is required.    
To this end, we use the the standard Procrustes method \cite{smith2017offline,Artetxe:2018acl} to align the embedding spaces of two languages, with bilingual dictionaries from \cite{glavas-etal-2019-properly} as the supervision guiding the alignment. Concretely, for each language pair in our experiments we project the AOC (ISO) embeddings of the source language to the AOC (ISO) space of the target language. 

\vspace{1.8mm}

\noindent \textbf{Direct Text Embedding with Multilingual Transformers}. 
%
%(we opt for proc over proc-B for simplicity). 
%
%Figure \ref{fig:contextfrequency} analyzes the impact of different thresholds for $\tau$.%
%
%In AOC we scan Wikipedia and collect for each term and for each containing sentence a contextual representation. The final representation is obtained by averaging over all collected embeddings. \todo{Formally?} In both ISO and AOC we end up with a static embedding space for each language which we use in the same way as with CLWE (\S\ref{sec:clwes}). Previous work has shown that pre-trained encoders exhibit already some lose language alignment which in direct application yield unsatisfactory results. Because of this we use procrustes for each language pair  \cite{xing2015normalized,smith2017offline} to map ISO/AOC embeddings from the query language space to the document language space, we use the bilingual dictionaries provided by \cite{glavas-etal-2019-properly} (we opt for proc over proc-B for simplicity). 
%
In both AOC and ISO, we use the multilingual (contextual) encoders to obtain the static embeddings for word types (i.e., terms): we can then leverage in exactly the same ad-hoc retrieval setup (\S\ref{sec:clwes}) in which CLWEs had previously been evaluated \cite{litschko2019evaluating}. In an arguably more straightforward approach, we also use pretrained multilingual Transformers (i.e., mBERT or XLM) to directly encode the whole input text (\textbf{SEMB}).  
%
%The previous two approaches create static word embedding tables by either ignoring context information or creating an averaged-context representation for each word type. 
%We evaluate another approach, which does not derive a static word embedding table but rather uses the encoder (i.e., again mBERT or XLM) to create sentence embeddings directly (\textbf{SEMB}). 
We encode the input text by averaging the contextualized representations of all terms in the text (we again compute the weighted average, where the terms' IDF scores are used as weights, see \S\ref{sec:clwes}). For SEMB, we take the contextualized representation of each term $t_i$ to be the contextualized representation of its first subword token, i.e., $\overrightarrow{t_i} = \psi\left(\{\overrightarrow{wp_{i,k}}\}^K_{k = 1}\right) = \overrightarrow{wp_{i,1}}.$\footnote{In our initial experiments taking the vector of the first term's subword consistently outperformed averaging vectors of all its subwords.} 

%Possible choices for the aggregation function $\psi()$ include average-pooling or max-pooling. Our preliminary experiments have suggested that the best aggregation strategy for SEMB is to represent each term within a sentence/document $t_i$ with its first constituent subword embedding, that is, it holds $\overrightarrow{t_i} = $. We then represent the entire document or sentence through aggregation as before (see \S\ref{sec:clwes} again).

%% (IV, removed, this does not add much to the discussion imo)
%%CLWEs are  limited to a specific vocabulary size because at some point the long tail word representations become random. Multilingual encoders can encode almost anything \todo{Mmmm, max length limit?}, which is potentially an advantage over CLWEs. Since we are interested in downstream applicability we do not pre-filter the vocabulary. 

\subsection{Specialized Multilingual Sentence Encoders}
\label{sec:sent-specialized-models}
Off-the-shelf pretrained multilingual Transformers such as mBERT and XLM have been shown to produce poor sentence embeddings yielding sub-par performance in unsupervised text similarity tasks; therefore, in order to be successful in semantic text (sentences or paragraph) comparisons, they first need to be fine-tuned on text matching (typically sentence matching) datasets \cite{reimers2020making-distil,cao2019multilingual,Zhao:2020acl}. %Consequently, a line of recent work has been investigating novel approaches focused on learning universal sentence representations. 
Such encoders \textit{specialized for semantic similarity} are supposed to encode sentence meaning more accurately, supporting tasks that require unsupervised (ad-hoc) semantic text matching. In contrast to mBERT and XLM, which contextualize (sub)word representations, these models directly produce a semantic embedding of the input text. We provide a brief overview of the models included in our comparative evaluation.

\vspace{1.4mm}
\noindent
\textbf{Language Agnostic SEntence Representations (LASER)} \cite{artetxe2019LASER} adopts a standard seque\-nce-to-sequence architecture typical for neural machine translation (MT). It is trained on 223M parallel sentences covering 93 languages. The encoder is a multi-layered bidirectional LSTM and the decoder is a single-layer unidirectional LSTM. The 1024-dimensional sentence embedding is produced by max-pooling over the outputs of encoder's last layer. The decoder then takes the sentence embedding as additional input as each decoding step. The decoder-to-encoder attention and language identifiers on the encoder side are deliberately omitted, so that all relevant information gets `crammed' into the fixed-sized sentence embedding produced by the encoder. In our experiments, we directly use the output of the encoder to represent both queries and documents.

%% In \cite{artetxe2019LASER} the authors propose one of the first framework for learning universal \textit{Language Agnostic SEntence Representations} (LASER). 

\vspace{1.4mm}
\noindent
\textbf{Multilingual Universal Sentence Encoder (m-USE)}
%\footnote{https://tfhub.dev/google/universal-sentence-encoder-multilingual/3}
is a general purpose sentence embedding model for transfer learning and semantic text retrieval tasks \cite{yang-etal-2020-multilingual-muse}. It relies on a standard dual-encoder neural framework \cite{chidambaram-etal-2019-learning-dual-encoder-framework,ijcai2019-746} with shared weights, trained in a multi-task setting with an additional translation bridging task. For more details, we refer the reader to the original work. There are two pretrained m-USE instances available -- we opt for the 3-layer Transformer encoder with average-pooling.  
%to directly produce a fixed-sized representations for queries and documents.

%%For a given instance $(s_i^I,s_i^R)$ of an input sentence $s_i^I$ and response (translation) sentence $s^R_i$ the following approximate conditional probability is maximized:
%% (IV, removed all this, too detailed)
%%\begin{equation}
%%    \Tilde{P}\left(s_i^R|s_i^I\right) = \frac{e^{\phi\left(s_i^I,s_i^R\right)}}{\sum^K_{j=1} e^{\phi\left(s_i^R,s_j^R\right)}}
%%    \label{eq:softmax}
%%\end{equation}
%
%%The function $\phi$ uses the shared encoder to create representations of $s^I_i$ and $s_i^R$ first and then compute the dot-product. Because computing the softmax over all instances is intractable the authors use other sentences in the same batch as negative examples. 

%% The model is trained in a multi-task setting (e.g., r, with a translation bridging task to train the model on 16 languages. The primary tasks are 1) question answer retrieval, where the dot-product similarity between a question and answer is maximized; 2) Natural Language Inference \cite{snli:emnlp2015}; 3) Translation Ranking, where embeddings of translation-pairs are tuned to have high dot-product similarity. 

\vspace{1.4mm}
\noindent
\textbf{Language-agnostic BERT Sentence Embeddings (LaBSE)} \cite{feng2020language-labse} is another neural dual-encoder framework, also trained with parallel data. Unlike in LASER and m-USE, where the encoders are trained from scratch on parallel data, LaBSE training starts from a pretrained mBERT instance (i.e., a 12-layer Transformer network pretrained on the concatenated corpora of 100+ languages). In addition to the multi-task training objective of m-USE, LaBSE additionally uses standard self-supervised objectives used in pretraining of mBERT and XLM: masked and translation language modelling (MLM and TLM, see \S\ref{sec:rw}). 
%to leverage large monolingual corpora. 
For further model details, we refer the reader to the original work.

%%The authors extend the loss in Equation \ref{eq:softmax} by adding a margin $m$ around positive (i.e., true) translation pairs \cite{ijcai2019-746}, $\phi$ is the cosine between the two embeddings. 
%
%%\begin{equation*}
%%    \mathcal{L} = - \frac{1}{N} \sum^N_{i=1} \frac{e^{\phi (x_i, y_i)-m}}{e^{\phi (x_i, x_i)-m}+\sum^N_{n=1,n\ne i}e^{\phi (x_i, y_n)}}
%%\end{equation*}
%
%%They emphasize that the loss is asymmetric and depends on whether the softmax is taken over source sentence embeddings or over target sentence embeddings. Following this insight they propose a combined loss which considers each direction. The encoder is a transformer \cite{vaswani2017attention} initialized with mBERT \cite{mbert} weights.

\vspace{1.4mm}
\noindent
\textbf{DISTIL} 
%\footnote{https://github.com/UKPLab/sentence-transformers}
\cite{reimers2020making-distil} is a teacher-student framework for injecting the knowledge obtained through specialization for semantic similarity from a specialized monolingual transformer (e.g., BERT) into a non-specialized multilingual transformer (e.g., mBERT). It first specializes for semantic similarity a monolingual (English) teacher encoder $M$ using the available semantic sentence-matching datasets for supervision. In the second, \textit{knowledge distillation} step a pretrained multilingual student encoder $\widehat{M}$ is trained to mimic the output of the teacher model. 
For a given batch of sentence-translation pairs $ \mathcal{B} = \{(s_j, t_j)\}$, the teacher-student distillation training minimizes the following loss: 
{
\footnotesize
\begin{equation*}
\mathcal{J}(\mathcal{B}) = \frac{1}{|\mathcal{B}|} \sum_{j\in \mathcal{B}} \left[\left(M(s_j)-\widehat{M}(s_j)\right)^2 + \left(M(s_j)-\widehat{M}(t_j)\right)^2\right].
\end{equation*}
}
\noindent The teacher model $M$ is Sentence-BERT \cite{reimers2019sentence}, BERT specialized for embedding sentence meaning on semantic text similarity \cite{cer-etal-2017-semeval} and natural language inference \cite{williams2018broad} datasets. The teacher network only encodes English sentences $s_i$. The student model $\widehat{M}$ is then trained to produce for both $s_j$ and $t_j$ the same representation that $M$ produces for $s_j$.  
%of the SBERT sentence embedding space is preserved, that is, $\hat{M}(s_j) \approx M(s_j)$, while also keeping representations of translation pairs close to each other, that is, $\hat{M}(t_j) \approx M(s_i)$. 
We benchmark different DISTIL models in our CLIR experiments, with the student $\widehat{M}$ initialized with different multilingual transformers.

\section{Experimental Setup}
\label{sec:experintalsetup}

\textbf{Evaluation Data.} We follow the experimental setup of Litschko et al.\ \cite{litschko2019evaluating}, and compare the models from \S\ref{s:mencoders} on language pairs comprising five languages: English (EN), German (DE), Italian (IT), Finnish (FI) and Russian (RU). For document-level retrieval we run experiments for the following nine language pairs: EN-\{FI, DE, IT, RU\}, DE-\{FI, IT, RU\}, FI-\{IT, RU\}. We use the 2003 portion of the CLEF benchmark \cite{braschler2003clef},\footnote{\url{http://catalog.elra.info/en-us/repository/browse/ELRA-E0008/}} with 60 queries per language pair. The document collection sizes are 17K (RU), 55K (FI), 158K (IT), and 295K (DE). For sentence-level retrieval, also following \cite{litschko2019evaluating}, for each language pair we sample from Europarl \cite{Koehn:2005} 1K source language sentences as queries and 100K target language sentences as the ``document collection''.\footnote{Russian is not included in Europarl and we therefore exclude it from sentence-level experiments. Further, since some multilingual encoders have not seen Finnish data in pretraining, we additionally report the results over a subset of language pairs that do not involve Finnish.}

\vspace{1.8mm}
\noindent
\textbf{Baseline Models.} In order to establish whether multilingual encoders outperform CLWEs in a fair comparison, we compare their performance against the strongest CLWE-based CLIR model from the recent comparative study \cite{litschko2019evaluating}, dubbed Proc-B. Proc-B induces a bilingual CLWE space from pretrained monolingual \textsc{fastText} embeddings\footnote{\url{https://fasttext.cc/docs/en/pretrained-vectors.html}} using the linear projection computed as the solution of the Procrustes problem given the dictionary of word-translation pairs. Compared to simple Procrustes mapping, Proc-B iteratively (1) augments the word translation dictionary by finding mutual nearest neighbours and (2) induces a new projection matrix using the augmented dictionary. The final bilingual CLWE space is then plugged into the CLIR model from \S\ref{sec:clwes}.   

Our document-level retrieval SEMB models do not get to see the whole document but only the first $128$ word-piece tokens. For a more direct comparison, we therefore additionally evaluate the Proc-B baseline (Proc-B\textsubscript{LEN}) which is exposed to exactly the same amount of document text as the multilingual XLM encoder (i.e., the leading document text corresponding to first $128$ word-piece tokens) 
%an additional baseline comparable to $\text{SEMB}_{\text{XLM}}$ in which we apply proc-B on the same document sub-sequence (words making up fist 128 word-pieces), i.e. proc-B gets to see the same text as $\text{SEMB}_{\text{XLM}}$. 
Finally, we compare CLIR models based on multilingual Transformers to  a baseline relying on machine translation baseline (MT-IR). In MT-IR, 1) we translate the query to the document language using Google Translate and then 2) perform monolingual retrieval using a standard Query Likelihood Model \cite{ponte1998language} with Dirichlet smoothing \cite{zhai2004study}. 

%prob-B starts with a small seed dictionary and performs multiple iterations. In each iteration the current mapping is used to find cross-lingual nearest neighbors, these word-pairs are then added to the seed dictionary. Larger dictionaries lead to better mappings in later iterations. 

\vspace{1.8mm}
\noindent
\textbf{Model Details.} For all multilingual encoders we experiment with different input sequence lengths: $64$, $128$, $256$ subword tokens. 
%For SEMB we found that the performance can be slightly improved if instead of averaging word-piece embeddings into word embeddings we take the first word-piece embedding. 
For AOC we collect (at most) $\tau=60$ contexts for each vocabulary term: for a term not present at all in Wikipedia, we fall back to the ISO embedding of that term. We also investigate the impact of $\tau$ in \S\ref{sec:discussion}.
%
%All models use special tokens which means we use $[CLS]$ and $[SEP]$ for BERT-based models and $\langle s\rangle$ and $\langle/s \rangle$ for XLM-based models. 
For purely self-supervised models (SEMB, ISO, AOC) we independently evaluate representations from different Transformer layers (cf. \S\ref{sec:discussion}). For comparability, for ISO and AOC -- methods that effectively induce static word embeddings using multilingual contextual encoders -- we opt for exactly the same term vocabularies used by the Proc-B baseline, namely the top 100K most frequent terms from respective monolingual fastText vocabularies. 
%We found computing and using ISO embeddings on the full corpus vocabulary introduces too much noise and leads to worse results. Computing AOC representations on the full vocabulary is very time-intensive and impractical. 
%
%Models specialized for sentence-level semantic similarity (\S\ref{sec:sent-specialized-models}) based on parallel data for supervision produce fixed-sized representations which we use directly for retrieval. 
We additionally experiment with three different instances of the DISTIL model: (i) $\text{DISTIL}_{\text{XLM-R}}$ initializes the student model with the pretrained XLM-R transformer \cite{conneau2019unsupervised}; $\text{DISTIL}_{\text{USE}}$ instantiates the student as the pretrained m-USE instance \cite{yang-etal-2020-multilingual-muse}; whereas $\text{DISTIL}_{\text{DistilmBERT}}$ distils the knowledge from the Sentence-BERT teacher into a multilingual version of DistilBERT \cite{sanh2019distilbert}, a 6-layer transformer pre-distilled from mBERT.\footnote{Working with mBERT directly instead of its distilled version led to similar scores, while increasing running times.} For SEMB models we scale embeddings of special tokens (sequence start and end tokens, e.g., \texttt{[CLS]} and \texttt{[SEP]} for mBERT) with the mean IDF value of input terms. 

%$DISTIL_{\text{XLM-R}}$ refers to \textit{xlm-r-100langs-bert-base-nli-mean-tokens}, $DISTIL_{\text{m-USE}}$ refers to \textit{distiluse-base-multilingual-cased}, $DISTIL_{\text{DistilBERT}}$ refers to \textit{distilbert-multilingual-nli-stsb-quora-ranking}.
\section{Results and Discussion} 

\subsection{Document-Level Cross-lingual Retrieval}
\begin{table*}[t!]
\centering
\caption{Document-level CLIR results (Mean Average Precision, MAP). \textbf{Bold}: best model for each language-pair. *: difference in performance w.r.t. Proc-B significant at $p = 0.05$, computed via paired two-tailed t-test with Bonferroni correction.}
\vspace{1mm}
\def\arraystretch{0.95}
{\scriptsize
{%\fontsize{8pt}{8pt}\selectfont
\begin{tabularx}{\linewidth}{l X X X X X X X X X X X} 
\toprule
 & EN-FI & EN-IT & EN-RU & EN-DE & DE-FI & DE-IT & DE-RU & FI-IT & FI-RU & AVG & w/o FI \\ \midrule
 \textit{Baselines} \\\midrule
MT-IR & .276 & \textbf{.428} & .383 & \textbf{.263} & \textbf{.332} & \textbf{.431} & .238 & \textbf{.406} & .261 & \textbf{.335} & \textbf{.349} \\
Proc-B & .258 & .265 & .166 & .288 & .294 & .230 & .155 & .151 & .136 & .216 & .227 \\
$\text{Proc-B}_{\text{LEN}}$ & .165 & .232 & .176 & .194 & .207 & .186 & .192 & .126 & .154 & .181 & .196 \\ \midrule
\multicolumn{12}{l}{\textit{Models based on multilingual Transformers}} \\ \midrule
$\text{SEMB}_{\text{XLM}}$ & .199* & .187* & .183 & .126* & .156* & .166* & .228 & .186* & .139 & .174 & .178 \\
$\text{SEMB}_{\text{mBERT}}$ & .145* & .146* & .167 & .107* & .151* & .116* & .149* & .117 & .128* & .136 & .137 \\ \cdashline{1-12}[.4pt/1pt]
$\text{AOC}_{\text{XLM}}$  & .168 & .261 & .208 & .206* & .183 & .190 & .162 & .123 & .099 & .178 & .206 \\ 
$\text{AOC}_{\text{mBERT}}$ & .172* & .209* & .167 & .193* & .131* & .143* & .143 & .104 & .132 & .155 & .171 \\ \cdashline{1-12}[.4pt/1pt]
$\text{ISO}_{\text{XLM}}$  & .058* & .159* & .050* & .096* & .026* & .077* & .035* & .050* & .055* & .067 & .083 \\ 
$\text{ISO}_{\text{mBERT}}$  & .075* & .209 & .096* & .157* & .061* & .107* & .025* & .051* & .014* & .088 & .119 \\ \midrule
\multicolumn{9}{l}{\textit{Similarity-specialized sentence encoders (with parallel data supervision)}} \\ \midrule
$\text{DISTIL}_{\text{XLM-R}}$  & .216 & .190* & .179 & .114* & .237 & .181 & .173 & .166 & .138 & .177 & .167 \\
$\text{DISTIL}_{\text{USE}}$ & .141* & .346* & .182 & .258 & .139* & .324* & .179 & .104 & .111 & .198 & .258 \\
$\text{DISTIL}_{\text{DistilmBERT}}$ & \textbf{.294} & .290* & \textbf{.313} & .247* & .300 & .267* & \textbf{.284} & .221* & \textbf{.302}* & .280 & .280 \\ \cdashline{1-12}[.4pt/1pt]
LaBSE & .180* & .175* & .128 & .059* & .178* & .160* & .113* & .126 & .149 & .141 & .127 \\
LASER & .142 & .134* & .076 & .046* & .163* & .140* & .065* & .144 & .107 & .113 & .094 \\
m-USE & .109* & .328* & .214 & .230* & .107* & .294* & .204 & .073 & .090 & .183 & .254 \\
\bottomrule
\end{tabularx}
}}
\label{tab:clefselfsup}
\vspace{-1.5mm}
\end{table*}

We show the performance (MAP) of multilingual encoders on document-level CLIR tasks in Table~\ref{tab:clefselfsup}.
The first main finding is that none of the self-supervised models (mBERT and XLM in ISO, AOC, and SEMB variants) outperforms the CLWE baseline Proc-B. However, the full Proc-B baseline has, unlike mBERT and XLM variants, been exposed to the full content of the documents. A fairer comparison, against Proc-B\textsubscript{LEN}, which has also been exposed only to the first $128$ tokens, reveals that SEMB and AOC variants come reasonably close, albeit still do not outperform Proc-B\textsubscript{LEN}. This suggests that the document-level retrieval could benefit from encoders able to encode longer portions of text, e.g., \cite{beltagy2020longformer,zaheer2020big}. For document-level CLIR, however, these models would first have to be ported to multilingual setups. While SEMB and AOC variants exhibit similar performance, ISO variants perform much worse. The direct comparison between ISO and AOC demonstrates the importance of contextual information and seemingly limited usability of multilingual encoders as word encoders, if no context is available. 

%This is in part because some relevance signals appear at the end of documents. They fall outside the span that SEMB models get to process. If we account for this and compare SEMB instead against $\text{proc-B}_\text{LEN=128}$ we see that if the document-length is not an issue $\text{SEMB}_\text{XLM}$ achieves similar results. This warrants further research on representation models for longer sequences in future work and their adaptation to multilingual setup \cite{beltagy2020longformer,zaheer2020big}. 

Similarity-specialized multilingual encoders, which rely on pretraining with parallel data, yield mixed results. Three models, $\text{DISTIL}_\text{DistilmBERT}$, $\text{DISTIL}_\text{USE}$ and m-USE, generally outperform the Proc-B baseline\footnote{As expected, m-USE and $\text{DISTIL}_\text{USE}$ perform poorly on language pairs involving Finnish, as they have not been trained on any Finnish data.} 
%setting new state-of-the-art results in unsupervised document retrieval
LASER is the only encoder trained on parallel data that does not beat the Proc-B baseline. We believe this is because (a) LASER's recurrent encoder provides text embeddings of lower quality than Transformer-based encoders of m-USE and DISTIL variants and (b) it has not been subdued to any self-supervised pretraining like DISTIL models. Even the best-performing CLIR model based on a multilingual encoder ($\text{DISTIL}_\text{DistilmBERT}$) overall falls behind the MT-based baseline (MT-IR). However, the performance of MT-IR crucially depends on the quality of MT for the concrete language pair: for language pairs with weaker MT (e.g., FI-RU, EN-FI, FI-RU, DE-RU), $\text{DISTIL}_\text{DistilmBERT}$ can substantially outperform MT-IR (e.g., 9 MAP points for FI-RU and DE-RU); the gap in favor of MT-IR is, as expected, largest for most similar language pairs, for which also the most reliable MT systems exist (EN-IT, EN-DE). In other words, the feasibility and robustness of a strong MT-IR CLIR model seems to diminish with more distant language pairs and lower-resource language pairs. We plan to investigate this conjecture in more detail in future work.

The variation in results with similarity-specialized sentence encoders indicates that: (a) despite their seemingly similar high-level architectures typically based on dual-encoder networks \cite{cer2018universal}, it is important to carefully choose a sentence encoder in document-level retrieval, and (b) there is an inherent mismatch between the granularity of information encoded by the current state-of-the-art text representation models and the document-level CLIR task.

%encodings than the  based on an recurrent architecture, which is generally thought to be  the only model that is not transformer-based falls behind. Pre-training models with parallel data lead to representations that yield strong results, however these models however still perform worse than MT-IR. Parallel data supervision models also have the short-coming of missing out relevance signals due to the sequence length constraint, we address this topic further in Section \ref{sec:discussion}. One further disadvantage of directly using the output of parallel data supervision models is that there is no IDF scaling of constituent word embeddings. IDF-scaling effectively removes high-frequent terms (stopwords). We experimented with running $\text{DISTIL}_\text{DistilmBERT}$ on text where stopwords are filtered a priori (MAP: 0.254). This concludes that stopwords provide important contextualization information for multilingual encoders. Similar to \cite{ijcai2019-746} we also  encoding and scoring constituent sentences independently (MAP: 0.277) and mean-pooling sentence embeddings (MAP: 0.267).

%%%%%%%%%%%%%%%%%%%%%%%%%%%%%%%%%%%%%%%%%%%%%%%%%%%%%%%%%%%%%%%%%%%
%%%%%%%%%%%%%%%%%%%%%%%%%%%%%%%%%%%%%%%%%%%%%%%%%%%%%%%%%%%%%%%%%%%

\subsection{Sentence-Level Cross-Lingual Retrieval}

\begin{table*}[!t]
\centering
\caption{Sentence-level CLIR results (MAP). \textbf{Bold}: best model for each language-pair. *: difference in performance with respect to Proc-B, significant at $p = 0.05$, computed via paired two-tailed t-test with Bonferroni correction.}
\vspace{1mm}
\def\arraystretch{0.93}
{\scriptsize
{%\fontsize{8pt}{8pt}\selectfont
\begin{tabularx}{\linewidth}{l X X X X X X X X}
\toprule
& EN-FI & EN-IT & EN-DE & DE-FI & DE-IT & FI-IT & AVG & w/o FI \\\midrule
\textit{Baselines} \\\midrule
MT-IR & .639 & .783 & .712 & .520 & .676 & .686 & .669 & .723 \\
Proc-B & .143 & .523 & .415 & .162 & .342 & .137 & .287 & .427 \\\midrule
\multicolumn{9}{l}{\textit{Models based on multilingual Transformers}}\\ \midrule
$\text{SEMB}_{\text{XLM}}$ & .309* & .677* & .465 & .391* & .495* & .346* & .447 & .545\\
$\text{SEMB}_{\text{mBERT}}$ & .199* & .570 & .355 & .231* & .481* & .353* & .365 & .469 \\ \cdashline{1-9}[.4pt/1pt]
$\text{AOC}_{\text{XLM}}$  & .099 & .527 & .274* & .102* & .282 & .070* & .226 & .361 \\
$\text{AOC}_{\text{mBERT}}$ & .095* & .433* & .274* & .088* & .230* & .059* & .197 & .312 \\ \cdashline{1-9}[.4pt/1pt]
$\text{ISO}_{\text{XLM}}$  & .016* & .178* & .053* & .006* & .017* & .002* & .045 & .082 \\ 
$\text{ISO}_{\text{mBERT}}$  & .010* & .141* & .087* & .005* & .017* & .000* & .043 & .082 \\ \midrule
\multicolumn{9}{l}{\textit{Similarity-specialized sentence encoders (with parallel data supervision)}} \\ \midrule
$\text{DISTIL}_{\text{XLM-R}}$ & .935* & .944* & .943* & .911* & .919* & .914* & .928 & .935 \\
$\text{DISTIL}_{\text{USE}}$ & .084* & .960* & .952* & .137 & .920* & .072* & .521 & .944 \\
$\text{DISTIL}_{\text{DistilmBERT}}$ & .817* & .902* & .902* & .810* & .842* & .793* & .844 & .882 \\ \cdashline{1-9}[.4pt/1pt]
LaBSE & .971* & .972* & .964* & .948* & .954* & .951* & .960 & .963 \\
LASER\qquad\qquad & \textbf{.974*} & \textbf{.976*} & \textbf{.969*} & \textbf{.967*} & \textbf{.965*} & \textbf{.961*} & \textbf{.969} & \textbf{.970} \\
m-USE & .079* & .951* & .929* & .086* & .886* & .039* & .495 & .922 \\ \bottomrule
\end{tabularx}
}}
\label{tab:europarl-selfsupervised}
\vspace{-1.5mm}
\end{table*}

We show the sentence-level CLIR performance in Table~\ref{tab:europarl-selfsupervised}. 
Unlike in the document-level CLIR task, self-supervised SEMB variants here manage to outperform Proc-B. 
%Especially SEMB-based models gain a big boost and outperform the proc-B baseline. $\text{SEMB}_{\text{XLM}}$ outperforms CLWE's, but still falls behind the machine translation baseline. 
The better relative SEMB performance than in document-level retrieval is somewhat expected: sentences are much shorter than documents (i.e., typically shorter than the maximal sequence length of $128$ word pieces). All purely self-supervised mBERT and XLM variants, however, perform worse than the translation-based baseline. 
%MT
%There is no problem with missing a relevance signal. At the same time the performance difference of AOC and ISO models, having no sequence-length constraint, are much closer to those achieved in document-level retrieval. Still all self-supervised models fall behind the MT-IR baseline. 

Multilingual encoders specialized with parallel data excel in sentence-level CLIR, all of them substantially outperforming the competitive MT-IR baseline. This however, does not come as much of a surprise, since these models (a) have been trained using parallel data, and (b) have been optimized exactly on the sentence similarity task. In other words, in the context of the cross-lingual sentence-level task, these models are effectively supervised models.  
%those models are trained directly the models are directly optimized for translation-pair matching when they are optimized for bitext retrieval.
The effect of supervision is most strongly pronounced for LASER, which was, by being also trained on parallel data from Europarl, effectively subdued to in-domain training. We note that at the same time LASER was the weakest model from this group on average in the document-level CLIR task. %It's still evident that while LASER exhibits the worst result among parallel data supervision models in document retrieval it performs very strong in sentence retrieval. This suggests that fine-tuning a model for translation performance only is insufficient for general purpose representations.

%%%%%%%%%%%%%%%%%%%%%%%%%%%%%%%%%%%%%%%%%%%%%%%%%%%%%%%%%%%%%%%%%%%
%%%%%%%%%%%%%%%%%%%%%%%%%%%%%%%%%%%%%%%%%%%%%%%%%%%%%%%%%%%%%%%%%%%

\subsection{Further Analysis}
\label{sec:discussion}
We further investigate three aspects that may impact CLIR performance of multilingual encoders: (1) layer(s) from which we take vector representations, (2) number of contexts used in AOC variants, and (3) sequence length in document-level CLIR.

\begin{figure}[!t]
\centering\includegraphics[width=\textwidth]{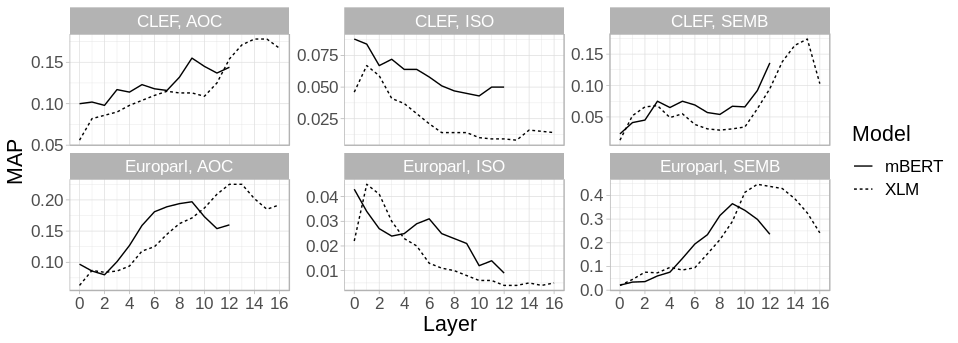}
\caption{CLIR performance of mBERT and XLM as a function of the Transformer layer from which we obtain the representations. Results (averaged over all language pairs) shown for all three encoding strategies (SEMB, AOC, ISO).}
\label{fig:layerplots}
\end{figure}

\vspace{1.8mm}
\noindent \textbf{Layer Selection.} All multilingual encoders have multiple layers and one may select (sub)word representations for CLIR at the output of any of them.    
%there are multiple ways to extract embeddings, when experimenting with self-supervised models we opt for treating and evaluating layers independently\footnote{We tried averaging layers bottom-up and top-down but could not achieve better results.}. 
Figure~\ref{fig:layerplots} shows the impact of taking subword representations after each layer for self-supervised mBERT and XLM variants. We find that the optimal layer differs across the encoding strategies (AOC, ISO, and SEMB) and tasks (document-level vs. sentence-level CLIR). ISO, where we feed the terms into encoders without any context, seems to do best if we take the representations from lowest layers. This makes intuitive sense, as the parameters of higher Transformer layers encode compositional rather than lexical semantics \cite{Ethayarajh:2019emnlp,rogers2020primer}. For AOC and SEMB, where both models obtain representations by contextualizing (sub)words in a sentence, we get the best performance for higher layers -- the optimal layers for document-level retrieval (L9/L12 for mBERT, and L15 for XLM) seem to be higher than for sentence-level retrieval (L9 for mBERT and L12/L11 for XLM). 
%Contrary to the representation strategies the best ISO embeddings are extracted from early layers, i.e. the embedding layer for mBERT and first layer for XLM. Logically, as context gets made available upper layers provide more meaningful representations. Moving from document-level to sentence-level retrieval we observe a shift of optimal layers towards earlier layers. The trend difference between ISO and SEMB/AOC suggests that lower layers encode more lexical word knowledge and upper layers encode more contextual/compositional knowledge.
%

\begin{figure}[!t]
    \centering
    \includegraphics[width=\textwidth]{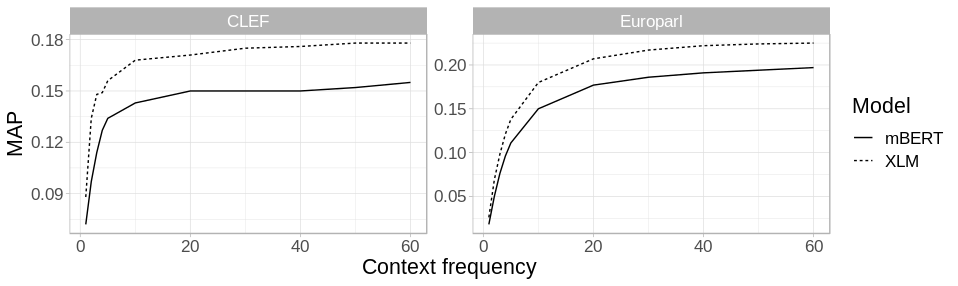}
    \caption{CLIR performance of AOC variants (mBERT and XLM) w.r.t. the number of contexts used to obtain the term embeddings.}
    \label{fig:contextplots}
    \vspace{-2mm}
\end{figure}

\vspace{1.8mm}
\noindent\textbf{Number of Contexts in AOC.} We construct AOC term embeddings by averaging contextualized representations of the same term obtained from different Wikipedia contexts. This raises an obvious question of a sufficient number of contexts needed for a reliable (static) term embedding. Figure~\ref{fig:contextplots} shows the AOC results depending on the number of contexts used to induce the term vectors (cf. $\tau$ in \S\ref{s:mencoders}). The AOC performance seems to plateau rather early -- at around 30 and 40 contexts for mBERT and XLM, respectively. Encoding more than 60 contexts (as we do in our main experiments) would therefore bring only negligible performance gains.  
%We can see that for both models on both datasets the performance levels off at a threshold between twenty and thirty embeddings. Considering more embeddings yields only marginal improvements and using fewer embeddings results in steep performance decreases.

\vspace{1.8mm}
\noindent\textbf{Input Sequence Length.} Multilingual encoders have a limited input length and they, unlike CLIR models operating on static embeddings (Proc-B, as well as our AOC and ISO variants), effectively truncate long documents. In our main experiments we truncated the documents to first $128$ word pieces. Now we quantify (Table~\ref{tbl:seqlen}) if and to which extent this has a detrimental effect on document-level CLIR performance.  
%text encoders compared to CLWE'S is that text encoders don't get to see the full document, due to the maximal sequence length constraint. This is because we cut out relevance signals appearing at the end of documents. 
Somewhat counterintuitively, encoding a longer chunk of documents ($256$ word pieces) yields a minor performance deterioration (compared to the length of $128$) for \textit{all} multilingual encoders. We suspect that this is a combination of two effects: (1) it is more difficult to semantically accurately encode a longer portion of text, leading to semantically less precise embeddings of $256$-token sequences; and (2) for documents in which the query-relevant content is not within the first $128$ tokens, that content might often also appear beyond the first $256$ tokens, rendering the increase in input length inconsequential to the recognition of such documents as relevant.    

%that simply increasing the length yields inferior results and is not the right strategy to address this problem. We hypothesize that simply including longer text span also introduces, in addition to the relevance signals, also more noise. Most models have an optimal sequence length at 128 while LASER being a BiLSTM-based model falls off in performance as sequence lengths get longer than 64. For both document retrieval and sentence retrieval we find that languages involving Finnish show low results. We posit that one of the reasons for this is related to the sequence length. Since Finnish contains the longest words, splitting words into many word-piece tokens has the effect that the model effectively gets to process shorter sentences. 

%\begin{figure}[!t]
%    \centering
%    \includegraphics[width=0.8\textwidth]{img/seqlenplot_v8.png}
%    \caption{Document CLIR results w.r.t. the input text length. Scores averaged over all language pairs not involving FI.}
%    \label{fig:seqlenplot}
%    \vspace{-0.5em}
%\end{figure}
\setlength{\tabcolsep}{2.5pt}
\begin{table*}[!t]
\centering
\caption{Document CLIR results w.r.t. the input text length. Scores averaged over all language pairs not involving Finnish.}
\vspace{1mm}
\def\arraystretch{0.99}
{\scriptsize
{%\fontsize{8pt}{8pt}\selectfont
\begin{tabularx}{\linewidth}{l X X X X X X X X}
\toprule
Length & $\text{SEMB}_\text{mBERT}$ & $\text{SEMB}_\text{XLM}$ & $\text{DIST}_\text{use}$ & $\text{DIST}_\text{XLM-R}$ & $\text{DIST}_\text{DmBERT}$ & mUSE & LaBSE & LASER \\ \midrule
64 & .104 & .128 & .235 & .167 & .237 & .254 & .127 & .089 \\
128 & .137 & .178 & .258 & .162 & .280 & .247 & .125 & .068 \\
256 & .117 & .158 & .230 & .146 & .250 & .197 & .096 & .027 \\ \bottomrule
\end{tabularx}
}}
\label{tbl:seqlen}
\end{table*}

\noindent 

% We hypothesize that word representations, as averaged word-piece embeddings, are likely to be less reliable for long words. The authors of \cite{glavas-etal-2019-properly} show that there is no one size fits all evaluation for CLWEs with respect to different cross-lingual transfer tasks, in this work we similarly show that retrieval performance is dependent on the type of retrieval and changes between sentence-level, document-level, and possibly word-level retrieval (BLI). 

%In summary we find that multilingual text encoders exhibit different degree of usefulness for unsupervised CLIR. Results vary between different levels of supervision. Self-supervised models generally underperform compared to CLWE's on document retrieval. If we account for the problem of sequence length we find that SEMB gets in the same ballpark as $\text{proc-B}_\text{LEN=128}$. Considering it's computational overhead this is an unsatisfactory result. Results are largely improved by moving to models with parallel supervision, setting new state-of-the-art results. On sentence retrieval both paradigms, self-supervision, parallel data models, yield strong results exceeding over the MT-IR baseline. Results are dependent on optimal layer selection and having collected a minimum number of contexts for AOC.

\section{Conclusion}

% Pretrained multilingual text encoders, based on neural Transformer architectures, such as multilingual BERT (mBERT) and XLM have achieved strong performance on a myriad of supervised language understanding tasks. Consequently, they have been adopted as a state-of-the-art paradigm for multilingual and cross-lingual representation learning and transfer, rendering cross-lingual word embeddings (CLWEs) effectively obsolete. However, questions remain to which extent this finding generalizes 1) to unsupervised settings and 2) for ad-hoc cross-lingual IR (CLIR) tasks. Therefore, in this work we present a systematic empirical study focused on the suitability of the state-of-the-art multilingual encoders for cross-lingual document and sentence retrieval tasks across a large number of language pairs. In contrast to supervised language understanding, our results indicate that for unsupervised document-level CLIR -- a setup in which there are no relevance judgments for task-specific fine-tuning -- the pretrained encoders fail to significantly outperform models based on CLWEs. For sentence-level CLIR, we demonstrate that state-of-the-art performance can be achieved. However, the peak performance is not met using the multilingual encoders `off-the-shelf', but rather relying on their variants that have been further specialized for sentence understanding tasks.

Pretrained multilingual (mostly Transformer-based) encoders have been shown to be widely useful in natural language understanding (NLU) tasks, when fine-tuned in supervised settings with some task-specific data; their utility as general-purpose text encoders in unsupervised multilingual settings, such as the ad-hoc cross-lingual IR, has been much less investigated. In this work, we systematically validated the suitability of a wide spectrum of cutting-edge multilingual encoders for document- and sentence-level CLIR across several language pairs. Our study encompassed purely self-supervised multilingual encoders, mBERT and XLM, as well as the multilingual encoders that have been specialized for semantic text matching on semantic similarity datasets and parallel data. Opposing the main findings from supervised NLU tasks, we have demonstrated that self-supervised multilingual encoders (mBERT and XLM), without exposure to any further supervision, in most settings fail to outperform CLIR models based on cross-lingual word embeddings (CLWEs). Semantically-specialized multilingual sentence encoders, on the other hand, do outperform CLWEs, but the gains are pronounced only in the sentence retrieval task. While state-of-the-art multilingual text encoders excel in so many seemingly more complex language understanding tasks, our work renders ad-hoc CLIR in general and document-level CLIR in particular a serious challenge for these models. We make our code and resources available at \url{https://github.com/rlitschk/EncoderCLIR}.

\bibliographystyle{splncs04}
\bibliography{references}

%\clearpage
%\appendix
%\input{xx-appendix.tex}

\end{document}